\documentclass[letterpaper]{article}
\usepackage{aaai}
\usepackage{times}
\usepackage{helvet}
\usepackage{courier}

\usepackage{svg}
\usepackage{times}
\usepackage{latexsym}
\usepackage{enumitem}
\usepackage{array,color}
\usepackage{multirow}
\usepackage{todonotes}
\usepackage{amsmath,amsfonts} % Math packages
\usepackage{tikz,amsmath,amsthm}
\usepackage{hyperref}
\usepackage{algorithmic,algorithm,varwidth,booktabs}

\theoremstyle{definition}

\newcommand{\m}{{\textsc{FactGen}}}

\newcolumntype{L}{>{\arraybackslash}m{14cm}}
\newcolumntype{S}{>{\arraybackslash}m{5cm}}
\newcolumntype{A}{>{\arraybackslash}m{7cm}}

\nocopyright

\begin{document}
\title{Fact-Enhanced Synthetic News Generation}
\author{Kai Shu$^{\ddagger}$\thanks{Equal contributions.},
        Yichuan Li$^\flat$$^*$,
        Kaize Ding$^{\dagger}$ and
        Huan Liu$^{\dagger}$\vspace{0.1in} \\
        $^{\ddagger}$Illinois Institute of Technology, Chicago, IL, USA\\
        $^\flat$Worcester Polytechnic Institute, Worcester, MA, USA\\
        $^{\dagger}$Arizona State University, Tempe, AZ, USA\\
        $^{\ddagger}$\texttt{kshu@iit.edu}, $^{\flat}$\texttt{yli29@wpi.edu}, $^{\dagger}$\texttt{\{kaize.ding, huan.liu\}@asu.edu} \\
        
}
\maketitle
\begin{abstract}
\begin{quote}
The advanced text generation methods have witnessed great success in text summarization, language translation, and synthetic news generation. However, these techniques can be abused
% \kz{have been abused} 
to generate disinformation and fake news. To better understand the potential threats of synthetic news, we develop a novel generation method {\m} to generate high-quality news content. 
The majority of existing text generation methods either afford limited supplementary information or lose consistency between the input and output which makes the synthetic news less trustworthy. 
To address these issues, {\m} retrieves external facts to enrich the output and reconstructs the input claim from the generated content to improve the consistency among the input and the output.  
Experiment results on real-world datasets demonstrate that the generated news contents of {\m} are consistent and contain rich facts. We also discuss an effective defending technique to identify these synthetic news pieces if {\m} was used to generate fake news.  
\end{quote}
\end{abstract}

\section{Introduction}
% Text generation has been an important task for Natural Language Processing (NLP). 
With the success of natural language process such as recurrent neural networks~(RNNs)~\cite{rnn} and self-attentive~(SA) language models~\cite{AttentionAllYouNeed,gpt2}, there has been a significant performance improvement in text generation applications, like document summarization~\cite{gehrmann2018bottomup}, machine translation~\cite{johnson2016googles} and  synthetic news generation~\cite{data2news}. 
% Its applications like . One important application of text generation is synthetic news generation~\cite{gpt2,grover}.

These advanced methods are able to generate readable and realistic text.
% with limited information~\kz{what do you mean by limited}. 
For example, one can use the generative adversarial network (GAN)~\cite{aghakhani2018detecting} or sequence-to-sequence (seq2seq) model~\cite{yang2019read} to generate human-like comments. One recent approach named Grover~\cite{grover} aims to generate news pieces based on multiple attributes such as headlines, authors, and website domains, showing some promising early results. 
% \hy{COMMENTS ABOUT PARA. 1 and 2: too much words on those background knowledge, however, they could be put into the related work part. Please briefly summarize the most related works here and throw out the your concerns (next paragraph).}

However, these methods can also be abused to generate and amplify disinformation and fake news.  For example, the machine generated fake review threaten business reputations\footnote{https://bit.ly/349tPW2}and virtual characters sends generated story to spread propaganda\footnote{https://bit.ly/36if2e9}. The wide dissemination of synthetic disinformation and fake news will bring new challenges to the news ecosystem. 
Therefore, it becomes critical to understand synthetic fake news for further achieving accurate detection.

\begin{table}[t]
    \centering
    \small
        \caption{Example claim and the beginning part of a news pieces from CNN/DailyMail dataset. The \textbf{black bold} sentence fragments are the consistent word and \textit{\color{red} Itatic red} fragment is the supplementary information.  }
        
    \resizebox{\columnwidth}{!}{\begin{tabular}{c|A}
    \toprule
         Claim & { \small iran nuke framework agreement should be judged on merits, not disinformation.}  \\
         \toprule
         Content & {\small The united states and its negotiating partners reached \textbf{a very strong framework agreement with iran} in \textit{\color{red}lausanne , switzerland , on thursday} that limits iran 's nuclear program in such a way as to effectively \textbf{block it from building a nuclear weapon}. The debate that has already begun since the announcement of the new framework will likely result in \textit{\color{red}more heat than light}. It will not be helped by the \textbf{gathering swirl of dubious assumptions and doubtful assertions.} }\\ 
         \bottomrule
         
    \end{tabular}
    }

    \label{tab:example}
\end{table}

In the real-world scenario, fake news deliberately imitates the writing styles of real news, which makes it hard to be identified by human and computational detection methods~\cite{shu2020combating}. Both fake and real news usually contain additional facts\footnote{We follow the definition of \texttt{fact} as, according to Oxford Dictionary, the information used as evidence or as part of news article.} that are consistent and supplementary to the news claims. 
For example, in Table~\ref{tab:example} 
, the news mainly focuses on \textit{framework agreement with Iran}, and provide additional facts like the location and time of the agreement.  To eventually identify synthetic disinformation, from an adversarial perspective, we attempt to build a powerful synthetic news generation model by closing the inherent \textit{factual} discrepancies between human and machine-generated text. 
Existing methods on generating synthetic news may fall short with the following limitations: (1) \textit{factual inconsistency}, indicating the generated news contradict or refute the news claims; and (2) \textit{factual scarcity}, meaning the generated news content may miss essential details to supplement the claim.  However,  directly using or fine-tuning language models does not help as it is non-trivial to enhance factual consistency and richness on a language model directly. 
Therefore, in this study, we aim to address the following challenges in synthetic news generation: (1) how to generate news content related to a given claim/context; and (2) how to ensure that the generated content contains supplemental fact information.

Our solution to these challenges results in a novel framework {\m}\footnote{The code is available at https://github.com/bigheiniu/FactGen}. (\underline{Fact}-Enhanced Synthetic News \underline{Gen}eration). {\m} consists of three major components: 
(1) \textit{Pseudo-Self-Attentive (PSA) Language Model}~\cite{ziegler2019encoderagnostic}, where the customized encoder deceptively injects source information (claims and external facts) into pre-trained decoder for the generation. 
The adapted deceptive injection mechanism can resolve the mismatch between the untrained encoder and the well-trained decoder;
% contains the pre-trained decoder and a randomly initialized encoder which simulates the self-attention expanding decoder' key-value pairs;
(2) \textit{Fact Retriever}, which heuristically retrieves the supplemental information from external fact corpus to provide more candidate facts
during generation; and
(3) \textit{Claim Reconstructor}, a randomly initialized masked language model~\cite{devlin2019bert} which enhances the output consistency by reconstructing the masked claim tokens from both the representation of the generated content and the unmasked claim tokens.
During training, the \textit{PSA Language Model} takes the news claim and the retrieved facts from the \textit{Fact Retriever} as input, then generates highly consistent news content by incorporating the \textit{Claim Reconstructor} into the generation process. In this way, the proposed framework can generate both fact-consistent and fact-enriched news content. To summarize, our main contributions are as follows:

\begin{itemize}
    \item We study a novel problem of fact-enhanced synthetic news  generation, which aims to generate consistent and fact-enriched news content.
    \item We propose a principled framework {\m} generates realistic synthetic news by retrieving external facts and reconstructing the input claim.
    \item We conduct experiments on real-world datasets using quantitative and qualitative metrics to demonstrate the effectiveness of {\m} for synthetic news generation and its defense. 
\end{itemize}

\section{Methodology}\label{sec: approaches}
\begin{figure}[!tb]
    \centering
    \includegraphics[width=\columnwidth]{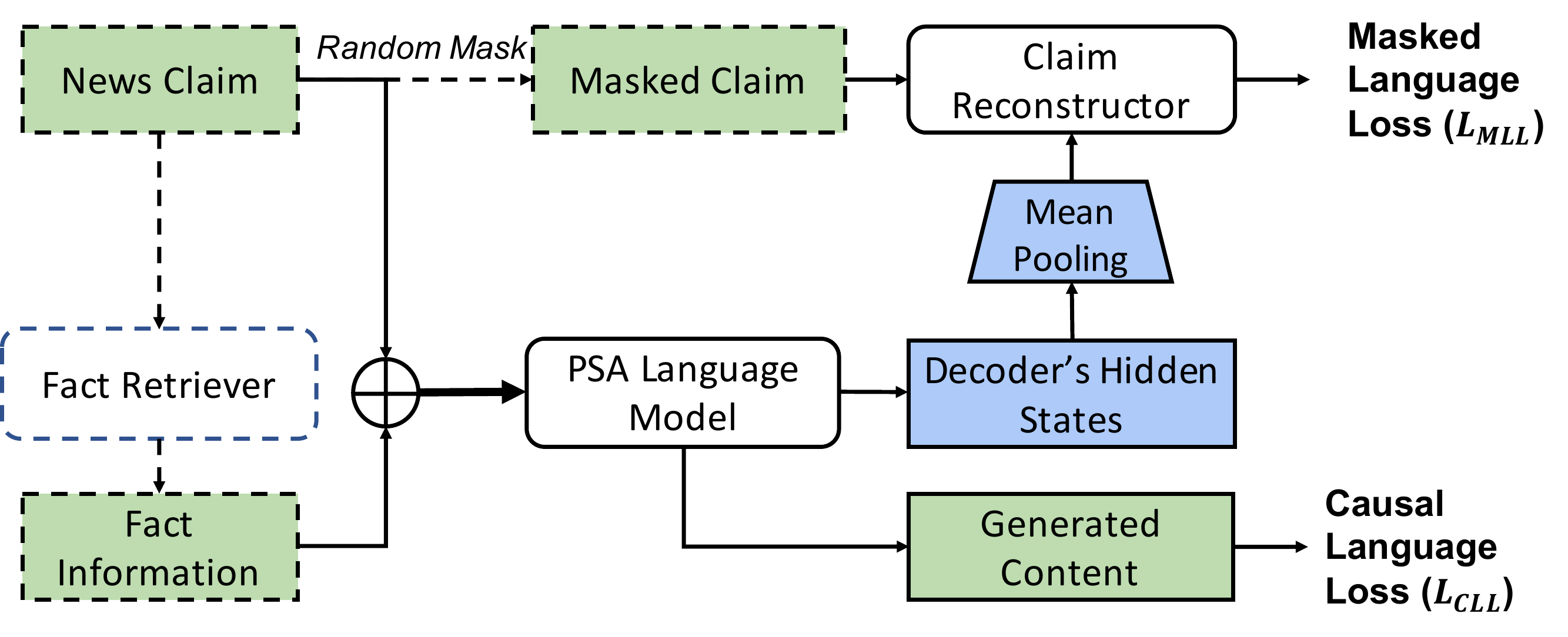}
    \caption{The proposed model, {\m}. The black dashed line indicates no differential dependency and the black bold line otherwise. $\oplus$ is the text concatenation. }
    % It is trained through minimizing the casual language loss and mask language loss. }
    % dependency.\hyli{Remember to re-plot this picture when you have time:1) The font need to be the same as your main paper (i.e., use times new roman); 2) the font size; 3) cut the blank area. }}
    \label{fig:model}
    %%%\vspace{-0.4cm}
\end{figure}

Our goal is to incorporate external facts into news generation that are consistent with the news claim. 
Given a sequence of tokens from the claim $X=\{x_1, x_2 \dots, x_N\}$, the fact retriever retrieves related fact information $F=\{f_1, f_2, \dots, f_K\}$ by semantic similarity, then the language model generates the news content $Y=\{y_1, y_2 \dots, y_M\}$ based on claim $X$ and $F$. It should be noticed that the length of $Y$ is much larger than $X$, which is $M >> N$, and $x_i, y_i, f_i$ are words.
Figure~\ref{fig:model} illustrates the architecture of our proposed model and the objective functions. The causal language loss $L_{CLL}$ depicts the loss of generating news content based on the input claim and fact.  The masked language loss $L_{MLL}$ is to reconstruct the masked input claim based on the language model output and the unmasked claims. 
This has a twofold benefit. Initially, the pre-trained decoder and the retrieved facts will bring unrelated information. This technique encourages the generated content to cover the input claim and provides a regularization effect. In addition, it is fully differentiable so we can minimize the objective function end-to-end. Overall, we minimize:
\begin{equation}
\small
    L = L_{CLL} + \lambda\;L_{MLL}
        % L = & \sum_{i=1}^{N}\sum_{t=1}^{m} -log
    % \;P(y_{t}|y_{1},\dots,y_{t-1};X_{i}, F_{i}) \\
    % & + \lambda\;\sum_{i=1}^{N}\sum_{x \in X_{[Masked]}} -log~P_i(x|X_{[No~Mask]},\; MeanPool(H_{Y}))
    \label{eq:objective}
\end{equation}
where $\lambda$ is the hyperparameter to control the contribution of claim reconstruction. The formulas of $L_{CLL}$ and $L_{MLL}$ are in Eq.~\ref{eq:CLL} and Eq.~\ref{eq:mll} respectively. 

\subsection{Preliminary}
Self-attentive~(SA) language models~\cite{devlin2018bert,gpt2,song2019mass} have achieved impressive performance gains in various language generation tasks.  These models are stacks of several SA blocks which encode the input $X=\{x_1, \dots, x_i, \dots, x_N\}$ into key-value pairs $(K, V) = \{(k_1, v_1), \dots, (k_i, v_i),\dots,(k_N, v_N)\}$ and  query $Q=\{q_1, \dots, q_i, \dots, q_N\}$. The next output is produced by taking the weighted sum of values $v_i$, where the weight assigned toward each value is the dot-product of the query $Q$ with all the keys $K$. The formula of SA is:
\begin{eqnarray}
K  = H_{X} W_k, ~~
    V  = H_{X} W_v,  ~~
    Q  = H_{X} W_q\\
    SA(X)  = \operatorname{softmax}\left(QK^{T}\right)V    
\end{eqnarray}
where $H_{X} \in \mathbb{R}^{N \times D} $ is the hidden representation of the input $X$, $D$ is the hidden dimension, and  $W_k, W_v, W_q \in \mathbb{R}^{D \times D}$ are the parameters to map the hidden representation of tokens $ H_{X}$ into key, value and query space, respectively.

\subsection{Proposed Method} 
\noindent\textbf{Pseudo-Self-Attentive Language Model:} 
Although the fine-tuned self-attentive language models like GPT-2~\cite{gpt2} have been applied to many text generation tasks, the application of using GPT-2 for the synthetic news generation may not be satisfactory. Since the GPT-2 is an autoregressive model, it will lose partial information from the input by only encoding the forward information. Besides, without a specific encoder, GPT-2 cannot capture the dependency relationship between the news claim and the retrieved facts which will hurt the performance of the decoder~\cite{edunov2019pretrained}. 
Therefore, we need a new encoder to capture bi-directional information and dependency among the input. 

\begin{figure}[tbp!]
    \centering
    \includegraphics[width=0.7\columnwidth]{ 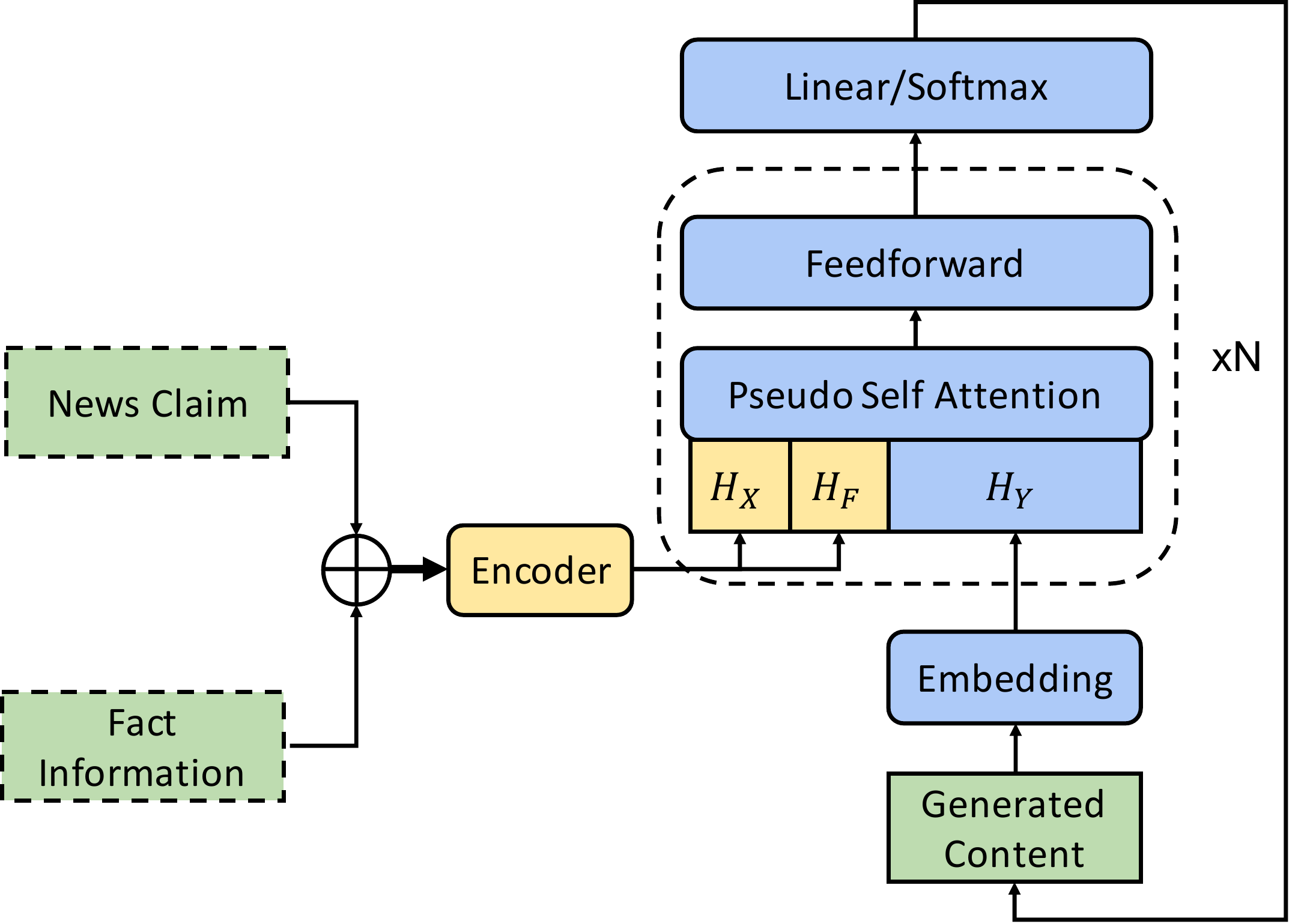}
    \caption{Our pseudo-self-attentive language model. Best visualized in color. The blue indicates decoder's pre-trained parameters. The yellow indicates the randomly initialized parameters of the encoder. $N$ is the number of PSA blocks.}
    \label{fig:model_LM}
    %%%\vspace{-0.4cm}
\end{figure}

We follow~\cite{ziegler2019encoderagnostic}'s setting, employing a pseudo-self-attentive(PSA) language model, where the "pseudo" is that the encoder deceptively extends the decoder's key-value pairs by the encoder's pairs, and the decoder predicts the next token not only based on previous output but also from the input. 
To model the dependency between claim and retrieved facts, we wrap them with ``[Claim]" and ``[Fact]" separately, and especially all the retrieved facts are contacted together without any special separation token. The architecture of the language model is shown in Figure~\ref{fig:model_LM} and  the formula of PSA is:
\begin{equation}
\small
\begin{split}
     \operatorname{PSA}(Y, X, F) =  
    \operatorname{softmax}\left(Q_{Y}\left[\begin{array}{c}
K_Y\\
K_X \\
K_F
\end{array}\right]^{\top}\right) 
 \left[\begin{array}{c}
V_Y \\
V_X \\
V_F
\end{array}\right]
\end{split}
\end{equation}
Note that $K_X, K_F, V_X$, and $ V_F$ are using different projection matrix $W_{*}$ and are randomly initialized. 
The objective function of the language model is:
\begin{equation}
\small
\label{eq:CLL}
    L_{CLL} = -\sum_{i=1}^{M} \left(log
    \;P(y_{i}|y_{1},\dots,y_{i-1};X, F) \right)
\end{equation}
% \kai{Please make sure you have correct and consistent notations. It is not clear why $W$ and $U$ is not in bold, and how these multiplication work. You can give the dimension for all the variables. Check whether $\textbf{H}=[\textbf{X}, \textbf{F}]$ or $\textbf{H}=[\textbf{X}; \textbf{F}]$}

\noindent\textbf{Fact Retriever:}
Directly training a sequence to sequence model on ${(X, Y)}$ often results in fact scarcity. One main reason is that facts from the input are extremely insufficient compared to the output. Thus, the language model is more likely to generate repeated sentences. Our solution towards the facts imbalance between the input and output is to increase the facts in the source side by retrieving related facts and considers them as part of the input.   
Our fact retriever~(FR) heuristically retrieves external facts in two steps. Firstly, to omit the computation limitation, we retrieve the related document based on the tf-idf vectors' cosine similarity between the claim and the document. Here we only keep the $top- k_1$ similar documents. Secondly, to accurately identify related sentences in the document, we utilize the pre-trained BERT~\cite{devlin2018bert} to encode all sentences presented in the picked documents and choose the $top-k_2$ most similar sentences based on the cosine similarity.

\noindent\textbf{Claim Reconstructor:}
% \todo{Compare with other methods that use back translation or classifiers. }
Since the aforementioned modules FR and PSA language model will bring inconsistency during the generation.
and high-quality news content is consistent with the news claims, to improve the content consistency, we propose to reconstruct the masked claim from the generated content and the unmasked claims. 

The existing reconstruction approaches for the input text consistence require the prior knowledge of the input, like  the topic label to learn a topic consistent reward function~\cite{yang-etal-2019-enhancing-topic}, key-entities for multi-classification on the hidden states to entail these entities in the generated content~\cite{wiseman2017challenges}.
Our claim reconstructor~(CR) does not require any prior knowledge about the input. It reconstructs the masked claim $X_{[Masked]}$ based on the mean pooling of output hidden representation $h_Y$ and unmasked sentence fragments $X_{[Unmasked]}$ .  
We mask claim's tokens 
with probability $P_{mask}$ and we follow the pseudo-self-attention~\cite{ziegler2019encoderagnostic} projecting $h_Y$ into CR's key-value pairs to predict the masked sentence fragment, $X_{[Masked]}$. The objective function of CR is:
\begin{equation}
\label{eq:mll}
\small
    L_{MLL} = \sum_{x \in X_{[Masked]}} -log~P(x|X_{[Unmasked]},\; h_{Y}))
%%%\vspace{-0.4cm}
\end{equation}

\begin{figure}
    \centering
    \includegraphics[width=0.76\columnwidth]{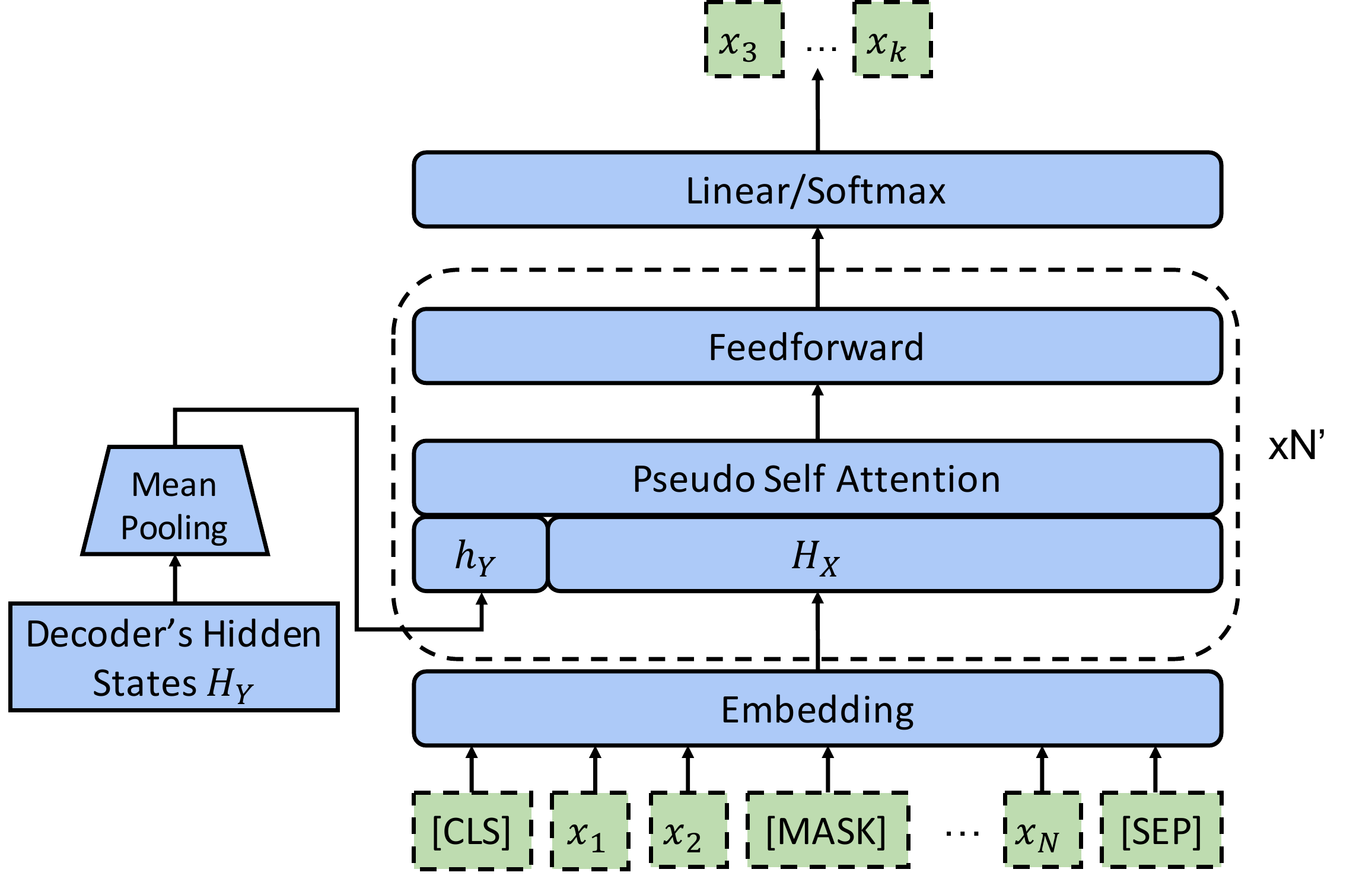}
    \caption{The overview of the claim reconstructor. It is also the PSA structure where we inject the mean pooling of decoder's hidden states to its key-value pairs. $N'$ is the number of PSA blocks.}
    \label{fig:cr}
    %%%\vspace{-0.4cm}
\end{figure}

\subsection{Training Schedule}
Since {\m} need to guarantee that two improvements will not contradict each other, we cannot directly train the model via minimizing eq~\ref{eq:objective}. We then train {\m} in two-stages. The overview of the training procedure is summarized in Algorithm~\ref{algo:training}. 
The joint training can align the latent space of these two modules. 
The two stages of training bring several advantages: firstly, it allows us to start the PSA language model and CR warmly, omitting the gradient explosion problem during training; secondly, because the claim is the main idea of the generated text and the retrieved facts are the auxiliary information during the generation, this order can help the decoder understand the importance of different input source. The following experiment result also stipulates the efficiency of this training schedule. 
\begin{algorithm}
  \caption{Training Procedure of {\m}}
  \begin{algorithmic}[1]
    \REQUIRE The source claims, relevant facts and target news pieces corpus $S=\{(X, F, Y)\}$; the masked and unmasked claims $D=\{(X_{[Masked]}, X_{[Unmasked]})\}$; first and second stage epoch number $epochs_1$ and $epochs_2$. 
    
    \ENSURE $PSA$ language model and claim reconstructor $CR$;
    
    \STATE Initialize $PSA_{encoder}$ and $CR$ with random weights
    \STATE Pre-train the $PSA$ via minimizing eq.\ref{eq:CLL} on $\{(X, Y)\}$; Pre-train the $CR$ via minimizing eq.\ref{eq:mll} on $D$.
    
   \FOR{$epoch=1$ to $epochs_1$} 
   \STATE Jointly training $PSA$ and $CR$ via minimizing eq.\ref{eq:objective} on $\{(X, Y)\}$ and $D$; ~~~\COMMENT{First Stage}
   \ENDFOR
    
    \FOR{$epoch=1$ to $epochs_2$}
   \STATE Jointly training $PSA$ and $CR$ via minimizing eq.\ref{eq:objective} on $S$ and $D$; ~~~\COMMENT{Second Stage}
   \ENDFOR

  \end{algorithmic}
  \label{algo:training}
\end{algorithm}

\section{Experiments}

% \hy{COMMENTS ABOUT THE EXPE. PART: the organization of this part need to be improved. For example, the ablation study could be put in another section, such as Analysis section. And I think putting the subsection ``Ablation Study'' before the ``Defending Synthetic Fake News"" makes me lost. Remember that the latter is your main contribution.}
In this section, we conduct experiments on two real-world news dataset to demonstrate the effectiveness of {\m} for news generation. Specifically, we aim to evaluate the quality of the generated news pieces in terms of \textit{fluency}, \textit{Consistency}, \textit{richness}, and \textit{trustworthiness}. 
% \textit{Fluency}: are the synthetic news contents fluent?
% \textbf{EQ2}: \textit{Consistency}: are the synthetic  news consistent with the claim?
% \textbf{EQ3}: \textit{Richness}: does synthetic news contain  rich facts?
% \textbf{EQ4}: \textit{Trustworthiness}: are synthetic news trustworthy?
% \textbf{EQ2}: \textit{Detectable}: are synthetic news easy to be detected by detection methods; \textbf{EQ3}: \textit{Defensible}: can our defending method detect these synthetic news contents?

% We aim to access the  by answering \textbf{EQ1} to \textbf{EQ4}, and evaluate the difficulty and our attempt of defending it through \textbf{EQ5} and \textbf{EQ6} respectively. 
\subsection{Dataset}
\label{sec:dataset}
We utilize two news dataset in our experiment.
The first dataset is a widely used fake news detection dataset collected from a fact-checking website, GossipCop~\cite{shu2018fakenewsnet}. Each sample contains the news' claim, content, metadata, label, and social engagements. The average lengths of the claim and content are 30 words and 250 words respectively.
The second dataset is the CNN/DailyMail news highlight dataset~\cite{hermann2015teaching} which contains the news content and selected highlight. In contrary to the text summarization, we use the highlight sentence as the source claim and the news content as the target text. On average, the claim has 56 tokens and the content has 790 tokens.
As for prepossessing, we truncate the news claim longer than 100 words and content longer than 300 words in both datasets. For the dataset splitting, we randomly sample 75\% training set, 15\% validation set, and 10\% test set in the GossipCop dataset and follow the same splitting setting in~\cite{see2017point}.  Datasets' statistical information is listed in Table ~\ref{tab:dataset_stat}.

\begin{table}[!ht]
\small
    \centering \caption{The statistical information of the datasets.}
    \begin{tabular}{c|c|c|c}
        \toprule
        Dataset & \# of train & \# of val & \# of test  \\
        \toprule
        GossipCop & 7,331 & 1,459 & 	974	 \\
        \hline
        CNN/DailyMail & 278,408 & 11,490 & 13,368\\
        \bottomrule
    \end{tabular}
    \label{tab:dataset_stat}
\end{table}

In this paper, we consider the factual sentences in the training dataset as our external fact corpus. This brings two advantages: firstly, utilizing several sentences instead of whole news pieces can avoid the model learning from copy the information from the source to the target side; secondly, the fact sentences from the training dataset can omit the data leakage problem during testing. Here we focus on external facts as the format of the text, though it can be extended to tabular data or knowledge graph.

\subsection{Experiment Settings}
We implement {\m} on  OpenNMT~\cite{klein-etal-2017-opennmt}. We tune the hyper-parameter $\lambda$ on the validation set. 
The encoder of {\m} is 4 blocks of SA block with 12 attention heads and 3072 hidden units. The weight of the decoder is initialized with the median pre-trained GPT-2~\cite{gpt2} model. The claim reconstruction module is 3 blocks of SA block with 4 attention heads and 256 hidden sizes. The optimizer is Adam~\cite{kingma2014adam} with $\beta_1 = 0.9$ and $\beta_2 = 0.998$. It should be noticed that the learning rate for the encoder is $1e-3$, for the decoder is $1e-5$, and $5e-5$ for the claim reconstruction. The number of retrieved documents ${k_1}$ and sentences ${k_2}$ is set to 10 and 5 respectively. The $epochs_1$ and $epochs_2$ in the training schedule are set to 4 and 2 respectively. 
During decoding we used Nucleus Sampling~(top-\textit{p}) with $p=0.9$. 
% \todo{how to set epoch1 and epochs2}

\subsection{Evaluation Metrics}
\subsubsection{Automatic Evaluation} 
The traditional text generation metrics like BLEU~\cite{bleu} and ROUGE~\cite{rouge} which are focus on the overlap between the generated content and the reference text which is not enough to reflect the claim-content consistency and the richness of the generated content. To remedy this, we develop two new evaluation metrics to measure the quality from different perspectives. 
\begin{itemize}
    \item \textbf{Fluency:} we report the BLEU score of different methods.
    % for reference
    \item \textbf{Consistency:} The ideal news content should support its claim. Therefore, we propose a stance detection model to detect whether the content is in favor of the claim or against it. 
    Given the claim and the generated news content $\{X, Y\}$, the stance detection model will output the relation of the text pair in \textit{(Agrees, Disagrees, Discusses, Unrelated)}.We utilize the Fake News Challenge dataset\footnote{http://www.fakenewschallenge.org/} to fine-tune  RoBERTa~\cite{liu2019roberta}.  This approach achieves a 0.93 accuracy score on the test dataset of the Fake News Challenge.
    We report the ratio of the ``agrees":
    \begin{equation}
    \small
    Consistency = \frac{\# of~agree~samples}{\#~of~all~samples}
    \end{equation}
    \item \textbf{Richness:} The richness of the output can be evaluated by the number of unique name entities in the generated text~\cite{fan2019strategies}. We utilize spaCy\footnote{https://spacy.io/} to extract the named entity from the output.

\end{itemize}

\noindent

\subsubsection{Human Evaluation} 
We distribute the 100 generated samples in CNN/DailyMail dataset to 2 annotators with a linguistic background. They have no advanced knowledge about the source of the generated content. They are asked to evaluate the generated content from fluency, richness, consistency, and Trustworthiness, 4 different perspectives\footnote{The details of human evaluation questions are in Appendix.}. So totally, there are 7,200 evaluation questions in our human evaluation. The annotator should answer each question from a score of 1 to 3 (3 being the best, 1 being the worst).
% \begin{itemize}
%     \item (Fluency) Is the generated news content readable? 
%     \item (Richness) Does the generated news content contain rich information?
%     \item (Consistency) Does the generated news content keep much fact information from the input.
%     \item (Trustworthy) Do you trust the information in the news content? 
% \end{itemize}{}

\subsection{Baseline Methods}
\label{sec:baselines}
To demonstrate the quality of the generated text, we compare our proposed model on content quality with the following text generation models: 
% \textbf{CopyTransformer}~\cite{see2017point}: a sequence-to-sequence transformer with a pointer network that can copy the word from the source to the target; \textbf{Conv Seq2Seq}~\cite{fan2018hierarchical}: it utilizes the seq2seq convolution neural network to generate claim consistent stories;  \textbf{PPLM}~\cite{dathathri2019plug}: a topic and content controlled language model; \textbf{GPT-2}~\cite{gpt2}: a large pre-trained language model which is the decoder part of the transformer. For a fair compairson with our model, we utilize the median size of the model; \textbf{Grover}~\cite{grover}: generating news text conditioned on the news title, authors, and website domains.
\begin{itemize}
    \item CopyTransformer~\cite{see2017point}: a sequence-to-sequence transformer with a pointer network that can copy the word from the source to the target. 
    \item Conv Seq2Seq~\cite{fan2018hierarchical}: a seq2seq convolution neural network to generate claim consistent stories.    
    \item PPLM~\cite{dathathri2019plug}: a topic and content controlled language model. It can directly control the pre-trained language model without fine-tuning.   
    \item GPT-2~\cite{gpt2}: a large pre-trained language model which is the decoder part of the transformer. For a fair comparison, we utilize the median size of the model.
    \item Grover~\cite{grover}: generating news text conditioned on the news title, authors, and website domains. It uses the same architecture as  GPT-2. 
\end{itemize}
\subsection{Experimental Results}
\begin{table*}[ht!]
\centering
\small
\caption{The performance comparison for the quality of the generated news pieces.}
\begin{tabular}{{lccccccc}}
\toprule
\multirow{2}{*}{Models}& \multicolumn{3}{c}{\centering {GossipCop}} & & \multicolumn{3}{c}{CNN/DailyMail} \\
 \cline{2-4}  \cline{6-8}
 & \multicolumn{1}{c}{BLEU} & \multicolumn{1}{c}{Richness} & \multicolumn{1}{c}{Consistency} & & \multicolumn{1}{c}{BLEU} & \multicolumn{1}{c}{Richness} & \multicolumn{1}{c}{Consistency} \\ 

CopyTransformer & 0.2 & 11.0 & 0.04 & & 0.5 & 9.5 & 0.66 \\
% InferEntityWriter & x & x & x & & x & x & x \\
ConvSeq2seq & 0.5 & 5.9 & 0.09 & &  3.3 &  9.5 & 0.44 \\
% SeqGAN & x & x & x & & x & x &  x \\
\midrule
PPLM & 0.7 & 12.5 & 0.67 & & 0.8 & 13.1 & 0.68 \\
GPT-2 & 0.8 & 13.4 & 0.35  & & 1.65 & 13.5 & 0.70 \\ 
Grover & 1.2 & 15.7 & 0.56  & & 0.3 & 15.3 & 0.72 \\ 
\midrule
\textbf{{\m}}  & \textbf{2.1} & \textbf{14.5} &  \textbf{0.80} & &\textbf{4.6} & \textbf{16.6} &  \textbf{0.76} 
\\ \bottomrule
\end{tabular} \label{tab:quality_result}
%%%\vspace{-0.3cm}
\end{table*}
\begin{table}[ht!]
% \small

\centering \caption{The human evaluation result of generated samples in the CNN/DailyMail dataset. We calculate the Pearson correlation to show the inter-annotator agreement.}
\small
\resizebox{\columnwidth}{!}{
\begin{tabular}{{lcccc}}
\toprule
% \multirow{2}{*}{Models}& \multicolumn{3}{c}{\centering {GossipCop}} & & \multicolumn{3}{c}{CNN/DailyMail} \\
 Methods & Fluency & Richness & Consistency & Trustworthiness  \\ 
\midrule
CopyTransformer & 1.68 & 1.65 & 1.89 & 1.62       \\
% InferEntityWriter & x & x & x & & x & x & x \\
ConvSeq2seq & 1.95 & 2.12 & 2.00 & 1.94 \\

% SeqGAN & x & x & x & & x & x &  x \\
\midrule
PPLM & 1.96  &  1.77 & 1.96 & 1.92 \\
GPT-2 & 2.03 & \textbf{2.32} & 1.95 & 2.08  \\ 
Grover & 2.08 & 2.15 & 1.78 & 1.97   \\ 
\midrule
\textbf{{\m}}  & \textbf{2.17} & 2.28 &  \textbf{2.12} &\textbf{2.18} \\ 
\midrule
Correlation & 0.14 & 0.26 & 0.21 & 0.21 
\\
% \textit{p-value} & 0.19 & 0.75 & 0.36 & 0.36 \\

\bottomrule
\end{tabular}} \label{tab:human_result}
%%%%\vspace{-0.4cm}
\end{table}
The automatic and human evaluation results are shown in Table~\ref{tab:quality_result} and \ref{tab:human_result}, respectively. We evaluate the quality of the text generation  through the following perspectives:
\begin{itemize}
    \item \textbf{Fluency:} From the human evaluations on fluency in CNN/DailyMail dataset and BLEU scores in two datasets, we can find that our model achieves the best performance. In the meantime, we find that the pre-trained language model achieves better human evaluation results than the model trained from scratch (PPLM, GPT-2, Grover $>$ CopyTransformer, Conv Seq2Seq). This indicates the importance of incorporating the large pre-trained language model in the synthetic news generation. Besides, {\m}'s performance indicates the pseudo-self attention properly connecting the randomly initialized encoder and the pre-trained decoder.
    \item \textbf{Consistency:} The consistent result in both human evaluation and automatic evaluation demonstrates the effectiveness of our approach. Especially, in the GossipCop dataset,  our approach achieves 42\% performance improvement over the best baseline in automatic metric, and in CNN/DailyMail, the human evaluation also shows that our approach achieves 6\% performance improvement compared with the best baseline method. The main reason for the increase is reconstructing the claim increases the coverage of the output on the input information. 
    \item\textbf{Richness:} Our approach achieves the best performance in CNN/DailyMail dataset and the second performance in the GossipCop dataset. The reason for the ordinary performance in  GossipCop is that the size of the candidate documents in GossipCop is much smaller than the CNN/DailyMail~(7,331 $<$ 278,408). The FR cannot retrieve enough related facts from the external corpus and CR will reject the inconsistent facts during generation. This indicates that FR can bring rich facts in generation.
    \item\textbf{Trustworthiness:} Human evaluation of the Trustworthiness of synthetic news content indicates that overall, {\m} can generate high-quality text content. This helps us to understand the difference between machine-generated news content and true news in the future.
    
\end{itemize}

\subsection{Case Study}
\noindent One case study of the generated samples is listed in Table~\ref{tab:case_study}. We only reveal the output from the model with pre-trained language models and we have several observations: 
    \textit{(i)} Our model mainly talks about the \textit{agreement of nuclear weapons in Iran} and includes the supplemental information about \textit{Iraq and UK's} action toward \textit{nuclear weapons}. This brings more context information about the news claims and makes the generated news more convincing. 
    \textit{(ii)} Although Grover mentions much additional factual information, it is unrelated to the \textit{nuclear agreement} with \textit{Iran}.    
    \textit{(iii)} The outputs of GPT-2 and PPLM mainly discuss the \textit{nuclear agreement} without supplemental information about the agreement.

% From this table, we can find that our model achieves the best performance across all metrics in CNN/DailyMail dataset and best performance in BLEU and Consistency in GossipCop dataset. The reason for lower Richness score compared with Grover in GossipCop is the small size of training data in GossipCop and our model cannot query enough related fact information. The improvements on Consistency and BLEU score across two datasets demonstrate the effectiveness of our approach.

\begin{table*}[!th]
\small
\caption{Samples of generated fake news with different fake news generation methods. The \textbf{black bold} sentence fragment is the consistent word and \textit{\color{red} Italic red} fragment is the supplementary information.}
\centering
% \kai{shorten the examples with dot..}}
\begin{tabular}{cL}
% {p{1cm} p{4cm}}
\toprule
Claim & iran framework agreement on nukes \dots opponents cast doubts. \dots uranium stockpile, centrifuges, implement rigorous inspections; \dots judged on merits, not disinformation.\\
\midrule
% {CopyTrans} & {
% try should centrifuges facts, but \textbf{iran framework and iran facts says}. its on rigorous \textbf{klass} he , on , \textbf{framework to merits} , \textbf{klass} strong is {\color{red}\textit{uranium}} klass the but says is centrifuges , is disinformation uranium \textbf{stockpile} and implement \textbf{opponents} klass to the  \textbf{opponents}, merits and centrifuges says doubts. \dots
% } \\
% \midrule
% {Conv Seq2Seq} & {
% The \textbf{iranian nuclear} program in iran will be the latest in a series of `` new steps" that would {\color{red}{\textit{help the country become the next u.s. country}}} to be the only country to be the {\color{red}{\textit{new territory of the united states}}}. The united states, iran, iran, the united states, iran, the united kingdom and the united states. \dots} \\ 
% \midrule
{PPLM} & {i think the agreement has been agreed on the whole of \textit{\color{red}American and nauru},and is a good first step towards an \textbf{agreement} on other issues. \dots cut its \textbf{stockpiles} and  implement some strict inspections. i think this is a good first step towards an \textbf{agreement}.i think \textit{\color{red}the deal is good but its not done yet and it still needs to be signed and accepted by america},and it will have to meet certain conditions for american to accept its deal on \textbf{nuclear}} \\ 
\midrule
{GPT-2} & { The \textbf{iran nuclear} program does not have a single day's notice of a nuclear crisis, and it does not have a single day's notice of a \textbf{nuclear weapon }program. \dots  Last week, the government of \textbf{iran} signed a {\color{red}framework agreement} that \textit{allows} the \textit{\color{red}permanent establishment of the nuclear program} at all levels \dots close its existing \textbf{enrichment plant}, \dots cooperation and the reduction of the \textbf{stockpile}.
} \\  
\midrule
{Grover} & {Unlike (allegedly) other \textit{\color{red} traitors who've betrayed the established foreign policy approach} to the Iranian nuclear development , in addition to \textit{\color{red}erasing himself from IMDB}, with \textit{\color{red}a stigmata, an air cannon}, \dots published his own book for reasons which, honestly, I'm not sure are entirely clear. The main key player in the \textbf{NPT negotiations}} \\ 
\midrule

{\m} & {billionaire businessman \textbf{fichard klass} has joined the \textit{\color{red} british establishment} to try to prevent the sale of \textbf{nukes} after \textbf{failing to address its facts} on how the \textbf{iran regime} used them to try to arm its \textbf{nuclear programme}. \textit{\color{red}the businessman, who once ran a world bank and was governor of iraq}, said the nukes would cut \textbf{uranium stockpile}, \textbf{implement a rigorous inspections} \dots and try to defend its nuclear program. while \textbf{opponents} said a \textit{\color{red} strong nuclear test would help persuade the uk to develop nuclear weapons}, \textbf{klass} said he hoped most sanctions would be \textbf{judged on merits}, \textbf{rather than disinformation} .}\\ 
\bottomrule
\end{tabular} \label{tab:case_study}
%%%%\vspace{-0.4cm}
\end{table*}

% \subsection{Hyper-parameter Analysis}
\subsection{Ablation Study}
\begin{figure}[!h]
    \centering
    \includegraphics[width=0.6\columnwidth]{ 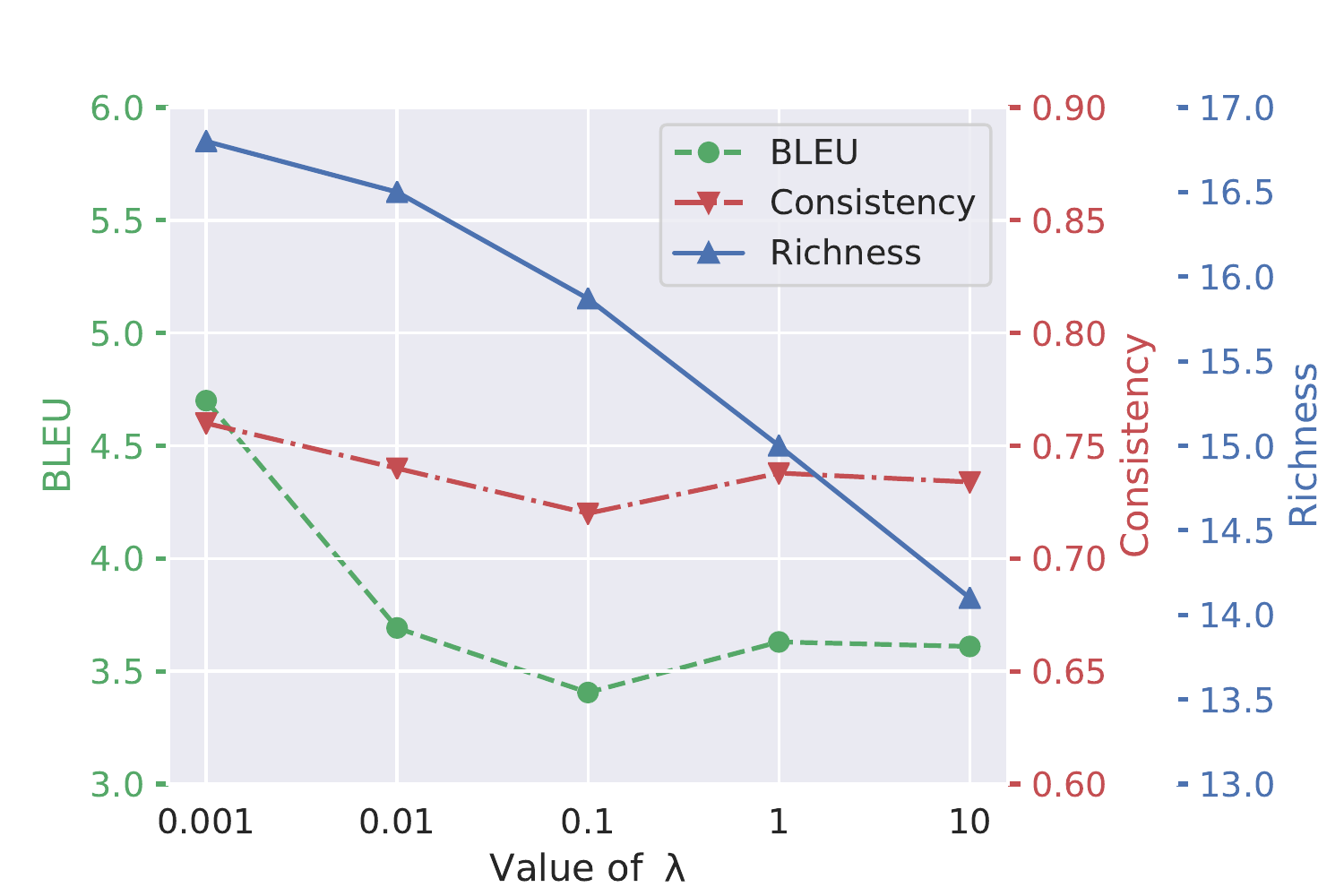}
    \caption{Impact of hyper-parameter $\lambda$ in CNN/DailyMail.}
    \label{fig:lambda}
\end{figure}

\noindent\textbf{Impact of $\mathbf{\lambda}$:} To learn the impact of the hyper-parameter $\lambda$ in our objective function in Eq.~\ref{eq:objective}, we change $\lambda$ from $\{0.001, 0.01, 0.1 , 1, 10\}$ and calculating all the automatic evaluation metrics. 
% All the models are trained  without FR in CNN/DailyMail dataset. 
From Figure~\ref{fig:lambda} we can find that $\lambda = 0.001$ achieves the best performance across all the automatic evaluations and with the increase of $\lambda$, the fact richness has been greatly decreased. This is because the CR will constrain the coverage of the generated content and cause the language model to only generated content around the input, which will reduce the richness of the generated content.

\noindent\textbf{Impact of Model Components:} To evaluate the importance of each key components, we set up three different ablation studies of {\m}: without claim reconstruction(\textit{w/o CR}), without Fact Retriever~(\textit{w/o FR}) and without these two components~(\textit{w/o CR and FR}). It should be noticed that all versions of the model have been pre-trained on $\{X, Y\}$. The automatic and human evaluation in Table~\ref{tab:ab_machine} and Table~\ref{tab:ab_human} show that the performance decrease in all ablation study. However, an interesting finding is that there seems to have a contradiction between the CR and FR. From Table~\ref{tab:ab_machine}, we find that \textit{w/o CR} contains the richest fact information but has the lowest consistency score; \textit{w/o FR} achieves the best BLEU score and compatible consistency score but the worst richness score. 
The impact of CR matches the observation of hyperparameter analysis, it improves the consistency of the generated content but has the side effects of decreasing the fact richness.
These results indicate the effectiveness of CR and FR in improving the richness and consistency in the generation.
\begin{table}[tbh!]
    \centering
    \small
    \caption{Results of automatic evaluation of model components ablation study in CNN/DailyMail dataset.}
    \resizebox{0.8\columnwidth}{!}{\begin{tabular}{cccc}
    \toprule
        Methods &  BLEU & Richness & Consistency \\
        \midrule
         Full Model & 4.7 & 16.8 & \textbf{0.76} \\
         \midrule
         \textit{single-stage} & 4.6 & 16.6 & 0.74 \\
         \textit{w/o CR} & 3.4 & \textbf{17.7} & 0.73 \\
         \textit{w/o FR} & \textbf{5.1} & 11.7 & 0.75 \\
         \textit{w/o CR and FR} & 4.0 &	12.9 & 0.73  \\
    \bottomrule
    \end{tabular}}
    
    \label{tab:ab_machine}
\end{table}

\begin{table}[tbh!]
    \centering
    \small
    \caption{ Results of human evaluation of model components ablation study in CNN/DailyMail dataset. }
    \resizebox{\columnwidth}{!}{\begin{tabular}{ccccc}
        \toprule
        Methods &  Fluency & Richness & Consistency & Trustworthiness \\
        \midrule
         Full Model & \textbf{2.17} & 2.28 & \textbf{2.12} & \textbf{2.18} \\
         \midrule
         \textit{single-stage} & 2.01 & 2.22 & 2.10 & 2.14 \\
         \textit{w/o CR} & 2.15 & \textbf{2.31} & 2.09 & 2.15 \\
         \textit{w/o FR} & 1.93 & 2.28 & 2.03 & 1.98 \\
         \textit{w/o CR and FR} & 2.09 & 2.19 & 2.03 & 2.10  \\
         \bottomrule
    \end{tabular}}
    
    \label{tab:ab_human}
\end{table}

\noindent\textbf{Impact of Training Schedule:}
To understand the effectiveness of our two-stage training schedule, we compare it with single-stage training where the model directly takes the claims and external fact information in the first stage. From the automatic and human evaluation result in Table~\ref{tab:ab_machine} and \ref{tab:ab_human}, we can find that two stages training schedule achieve better performance in all categories compared with single-stage. This stipulates the effectiveness of our training schedule. 

\section{Further Analysis}
\subsection{Difficulty of Defending Synthetic Fake News}
To understand the difficulty in synthetic fake news detection, we test the fake news detection methods and synthetic generation detection method on generated fake news and human-written real text. To guarantee the veracity of the test content, we select the fake generated content which is conditioned on fake claim and human-written real text is from real news pieces in GossipCop.   
To understand the difficulty in synthetic fake news detection, we test the fake news detection methods and synthetic generation detection method on generated fake news and human-written real text. To guarantee the veracity of the test content, we select the fake generated content which is conditioned on fake claims and human-written real text is from real news pieces in GossipCop.   
The reason for different training datasets for these approaches is to test whether the fake news detection model can transfer the knowledge in human written fake news into machine-generated fake news. To give limited access to generated content, the training dataset for both approaches will include extra 100 fake synthetic news pieces. 
% we increase the training dataset for both approaches by 100 generated fake news pieces. 
% 
% To omit the data leakage problem for further experiments, the randomly selected news pieces are from the test dataset in synthetic generation evaluation. 
% 
We test the classification accuracy in 300 fake generated news contents and the same amount of human-written real text. To omit the data leakage problem for evaluation, the test dataset is also the test data for synthetic generation evaluation. 
From the result in Table~\ref{tab:ex_defend}, we observe 
that fake news detection methods achieve worse performance than neural text classification (RoBERTAa $>$ EANN, MWSS-CNN) which indicates the difficulty of the current fake news detection method in detecting fake synthetic news. 
\subsection{Defending Against Synthetic Fake News}
To detect the new synthetic fake news, we follow~\cite{zellers2019defending} develop a defending method ${\m}_{def}$ based on the checkpoint of {\m} at iteration 20k. This setting can reduce the parameters overlap between the generator and the discriminator. We also use $h_Y$ as the final representation of the input, synthetic fake news or human written real news, and add a full connection layer to classify whether the input is fake or real. We utilize 300 synthetic fake news content and the same amount of human written real news to fine-tune ${\m}_{def}$. 
% RoBERTa, training to classify the human written and machine-generated text. 
The result in Table~\ref{tab:ex_defend} shows ${\m}_{def}$ achieves the best accuracy score. This is because ${\m}_{def}$ can learn a better representation of the input. We thus conclude that while the synthetic content is hard to be identified by existing methods, it still can be detected by ${\m}_{def}$. 

\begin{table}[tbh!]
    % \centering
        \small
        
        \caption{Results of synthetic fake news content detection.}
        \resizebox{\columnwidth}{!}{
    \begin{tabular}{ccc|c|c}
    \toprule
          &   EANN & MWSS-CNN & RoBERTa & ${\m}_{def}$ \\
          \midrule
          Accuracy & 0.64 & 0.58 & 0.74 & \textbf{0.82} \\
    \bottomrule
    \end{tabular}}
    \label{tab:ex_defend}
    
    %%%\vspace{-0.4cm}
    
\end{table}

\section{Related Work}
\iffalse 
2. style transfer
3. fake news detection
\fi
\noindent{\textbf{Synthetic News Generation}}
 Most synthetic news generation systems used in the newsroom are heavily rule-based and template-based~\cite{data2news}. The neural synthetic news generation like Grover~\cite{grover} utilizes an autoregressive language model to learns the dependency among news metadata fields include the domain, date, authors, title, and body. Sam and {et. al}~\cite{wiseman2017challenges} propose a structured data to text challenge which is to generate a sport a news piece of sports games from the associated box- or line-score data. To better capture the input data, \cite{wiseman2017challenges} employs copy-mechanism and source reconstruction as their seq2seq model extensions and \cite{puduppully2018datatotext}  generate text in recording plan and realization two stages.

\noindent{\textbf{Synthetic/Fake News Detection}}
Content-based fake news detection methods often leverage features from the feature engineering or latent features extracted by deep neural network~\cite{prezrosas2017automatic}. The deep learning models utilize the linguistic representation of news content to detect fake news. Qian \textit{et al.}~\cite{tcnnurg}  proposes a method learning the representation of news content and reconstructing the users comment during training, and in inference, this model makes a classification based on the representation of news content and the generated news comment for early fake news detection.  
Tal Schuster~\cite{schuster2020limitations} stipulates that current synthetic disinformation detection methods are mainly based on the stylometry which is limited against machine-generated misinformation.
% Dirk Hovy \textit{et al.} proposes an adversarial setting in detecting the generated reviews~\cite{hovy2016enemy}. 
 Gehrmann \textit{et al.} ~\cite{gehrmann2019gltr} visualize the distribution of every word that help the non-expert users find out the generated text. \cite{grover} and~\cite{solaiman2019release} propose neural generation detectors that fine-tune classifiers on the generator's previous checkpoint.  \cite{uchendu-etal-2020-authorship} try to identifying the NLP method of the generated text.

\section{Conclusion and Future Work}
In this paper, we propose a synthetic news generation method {\m}  to ensure fact-consistency and fact-richness. From the automatic and human evaluation of the content quality, {\m} is more effective than existing methods. Simultaneously, we discuss the difficulty of detecting synthetic fake news content by current SOTA fake news and human-machine detection methods. For social good, we propose a defending method ${\m}_{def}$ that achieves outstanding performance in detecting synthetic fake news content. In the future, we would like to include other formats of facts like tabular or knowledge graph. This can help us retrieve up-to-date fact information during generation. Since fake news often contains propaganda and more likely to widely spread on the social network, we would like to explore the style control of the generated content to make it prone to be spread.

\bibliographystyle{aaai}
\bibliography{cit}
\section{Appendices on Reproducibility}
In this section, we provide more details about the human evaluation questions, experimental settings and hyperparameter configuration to enable the  reputability of our work. 
\label{sec:appendix}
\subsection{Human Evaluation Question}
To evaluate the quality of {\m}, we ask human workers to answer four different questions. For each question, human worker need to give a score from 1 to 3 (1 means low quality and 3 is high quality).

\begin{itemize}

    \item ({Fluency}) Is the output article written by human?
    \item ({Richness}) Does the text provide extra information not listed in the input claim?
    \item (Consistency) Is the output consistent with the input? 
    \item ({Trustworthiness}) Do you trust the output content?
    
\end{itemize}

\subsection{Synthetic News Generation}
In Section~\ref{sec:baselines}, we compare {\m} with 5 baseline methods, including Conv Seq2Seq, CopyTransformer, PPLM, fine-tuned GPT-2 and Grover. 

For the dataset, GossipCop is available in the dataset section of the submission and CNN/DailyMail is available at \footnote{https://github.com/harvardnlp/sent-summary}. The description of the Fact Retriever as follows:
\begin{itemize}
    \item $top-{k_1}$: the number of documents we retrieved based on the tf-idf cosine similarity, we rank it from the biggest to the smallest.  We set $top-{k_1}$ to 10 for both datasets. 
    \item $top-{k_2}$: the number of sentences we retrieved from the $top-{k_1}$ Documents. We set it 5 for both datasets. 
\end{itemize}
% % \begin{table}[!h]
% %     \centering
% %     \caption{Parameter of Fact Retriever}
% %     \begin{tabular}{c|c|c}
% %         \toprule
% %         Parameter & GossipCop & CNN/DailyMail \\
% %         \midrule
% %         $top_{k_1}$ &  10 & 10 \\
% %         \midrule
% %         $top_{k_2}$ & 5 & 5  \\
% %         \bottomrule
% %     \end{tabular}
    
%     \label{tab:parameter}
% \end{table}
The parameters for evaluating the semantic similarly are the same as RoBERTa. 

\subsection{ Synthetic News Detection}
The two fake news detection models EANN, MWSS can be obtained online. \textit{EANN}: it is publicly available at https://github.com/yaqingwang/EANN-KDD18. \textit{MWSS-CNN}: it is avaliable at https://github.com/microsoft/MWSS.

The hyper parameters for human written and machine generated content detection RoBERTa are as follows:
\begin{itemize}
    \item Epochs: fine-tuning Epochs the RoBERTa model, 10. 
    \item Patience: the number of epochs to wait before early stop if no progress on the validation set, 3.
    \item Batch size: number of samples in one iteration, 10.
    \item Learning Rate: model fine-tuning learning rate, 5e-5.
    \item Max Length: we pad the input sentence into 300 tokens.
\end{itemize}
% \begin{table}[!h]
%     \centering
%     \caption{Hyper Parameters for RoBERTa}
%     \begin{tabular}{c|c}
%     \toprule
%         Parameters & Value \\
%     \midrule
%         Epochs &  10 \\
%         \midrule
%         Patience & 3 \\
%         \midrule
%         Batch Size & 10 \\
%         \midrule
%         Learning Rate & 5e-5\\
%         \midrule
%         Max Length & 300 \\
%         \bottomrule
%     \end{tabular}
    
%     \label{tab:roberta}
% \end{table}
\section{Appendices on Ethics Statement}
To better understand the characteristics of synthetic fake news, we propose a fact-enriched synthetic news generation method to generate high quality news pieces. From the automatic and human evaluation results, we find that {\m} can generate human-like and convincing news pieces. In this paper, we also discuss a possible solution to defend this attack, which is to use the checkpoint of {\m}. We are discussing the further usage of {\m} and ethical concerns as follows: 

\noindent{\textbf{Journalism Assistants:}} Since our method retrieves the external fact information and generate fact-consistent and fact-enriched news pieces, the journalists can utilize {\m} to automatically generate news by providing additional factual information and the claim. However, it still needs manually checking~\cite{leppanen-etal-2017-data}. 

\noindent{\textbf{Synthetic Disinformation Detection:}} In this paper, we shortly discuss the defending method, ${\m}_def$, and prove the effectiveness of it. However, like the Grover~\cite{grover}, this method mainly relies on semantic information rather than the veracity of the information~\cite{schuster2019limitations}. Future work should verify the factual correctness of the text in the following pipeline: check-worthy sentence extraction, the verified claim matching, and prediction~\cite{AdairAutomatedPF}.

\noindent\textbf{Release Policy:} Since {\m} can generate human-like and convincing news content, we need to critically release the code and the model parameters. We propose to publicly release the code including generator and discriminator. However, as for the checkpoints of both models, we will only share for academic usage.

\end{document}